\documentclass[sigconf, nonacm]{acmart}

\usepackage{algorithm}
\usepackage{algpseudocode}
\usepackage{booktabs}
\usepackage{multirow}
\usepackage{mathrsfs}
\usepackage{xcolor}
\usepackage{caption}
\usepackage{tcolorbox}
\usepackage{bbm}

\fontsize{10pt}{12pt}\selectfont

\newcommand\vldbyear{2024}
\newcommand\vldbworkshop{LLM+KG}
\newcommand\vldbauthors{\authors}
\newcommand\vldbtitle{\shorttitle}
\newcommand\vldbavailabilityurl{URL_TO_YOUR_ARTIFACTS}
\newcommand\vldbpagestyle{plain}

\begin{document}
\title{OneEdit: A Neural-Symbolic Collaboratively\\ Knowledge Editing System}

%%
%% The "author" command and its associated commands are used to define the authors and their affiliations.
\author{Ningyu Zhang}
\affiliation{%
  \institution{Zhejiang University}
  \city{Hangzhou}
  \state{Zhejiang}
  \country{China}
}
\email{zhangningyu@zju.edu.cn}

\author{Zekun Xi}
\affiliation{%
  \institution{Zhejiang University}
  \city{Hangzhou}
  \state{Zhejiang}
  \country{China}
}
\email{22351325@zju.edu.cn}

\author{Yujie Luo}
\affiliation{%
  \institution{Zhejiang University}
  \city{Hangzhou}
  \state{Zhejiang}
  \country{China}
}
\email{luo.yj@zju.edu.cn}

\author{Peng Wang}
\affiliation{%
  \institution{Zhejiang University}
  \city{Hangzhou}
  \state{Zhejiang}
  \country{China}
}

\author{Bozhong Tian}
\affiliation{%
  \institution{Zhejiang University}
  \city{Hangzhou}
  \state{Zhejiang}
  \country{China}
}

\author{Yunzhi Yao}
\affiliation{%
  \institution{Zhejiang University}
  \city{Hangzhou}
  \state{Zhejiang}
  \country{China}
}

\author{Jintian Zhang}
\affiliation{%
  \institution{Zhejiang University}
  \city{Hangzhou}
  \state{Zhejiang}
  \country{China}
}

\author{Shumin Deng}
\affiliation{%
  \institution{National University of Singapore, NUS-NCS Joint Lab}
  \city{Singapore}
  \country{Singapore}
}

\author{Mengshu Sun}
\affiliation{%
  \institution{Ant Group}
  \city{Hangzhou}
  \state{Zhejiang}
  \country{China}
}

\author{Lei Liang}
\affiliation{%
  \institution{Ant Group}
  \city{Hangzhou}
  \state{Zhejiang}
  \country{China}
}

\author{Zhiqiang Zhang}
\affiliation{%
  \institution{Ant Group}
  \city{Hangzhou}
  \state{Zhejiang}
  \country{China}
}

\author{Xiaowei Zhu}
\affiliation{%
  \institution{Ant Group}
  \city{Hangzhou}
  \state{Zhejiang}
  \country{China}
}

\author{Jun Zhou}
\affiliation{%
  \institution{Ant Group}
  \city{Hangzhou}
  \state{Zhejiang}
  \country{China}
}

\author{Huajun Chen}
\authornote{Corresponding author.}
\affiliation{%
  \institution{Zhejiang University}
  \city{Hangzhou}
  \state{Zhejiang}
  \country{China}
}


\begin{abstract}
Knowledge representation has been a central aim of AI since its inception. Symbolic Knowledge Graphs (KGs) and neural Large Language Models (LLMs) can both represent knowledge. KGs provide highly accurate and explicit knowledge representation, but face scalability issue; while LLMs offer expansive coverage of knowledge, but incur significant training costs and struggle with precise and reliable knowledge manipulation. To this end, we introduce \textbf{OneEdit}, a neural-symbolic prototype system for collaborative knowledge editing using natural language, which facilitates easy-to-use knowledge management with KG and LLM. OneEdit consists of three modules: 1) The \textit{Interpreter} serves for user interaction with natural language; 2) The \textit{Controller} manages editing requests from various users, leveraging the KG with rollbacks to handle knowledge conflicts and prevent toxic knowledge attacks; 3) The \textit{Editor} utilizes the knowledge from the Controller to edit KG and LLM.
We conduct experiments on two new datasets with KGs  which demonstrate that OneEdit can achieve superior performance.
%The results demonstrate that OneEdit, combined with existing knowledge editing techniques, achieves superior performance.
%We release an online system at \url{http://oneedit.zjukg.cn/}.
%Experimental results demosntrate that OneEdit can yield better performance than previous baselines for collaboratively knowledge editing. 
\end{abstract}

\maketitle

\pagestyle{\vldbpagestyle}
\begingroup\small\noindent\raggedright\textbf{VLDB Workshop Reference Format:}\\
\vldbauthors. \vldbtitle. VLDB \vldbyear\ Workshop: \vldbworkshop.\\ %\vldbvolume(\vldbissue): \vldbpages, \vldbyear.\\
%\href{https://doi.org/\vldbdoi}{doi:\vldbdoi}
\endgroup
\begingroup
\renewcommand\thefootnote{}\footnote{\noindent
This work is licensed under the Creative Commons BY-NC-ND 4.0 International License. Visit \url{https://creativecommons.org/licenses/by-nc-nd/4.0/} to view a copy of this license. For any use beyond those covered by this license, obtain permission by emailing \href{mailto:info@vldb.org}{info@vldb.org}. Copyright is held by the owner/author(s). Publication rights licensed to the VLDB Endowment. \\
\raggedright Proceedings of the VLDB Endowment. %, Vol. \vldbvolume, No. \vldbissue\ %
ISSN 2150-8097. \\
%\href{https://doi.org/\vldbdoi}{doi:\vldbdoi} \\
}\addtocounter{footnote}{-1}\endgroup

\ifdefempty{\vldbavailabilityurl}{}{
\vspace{.3cm}
\begingroup\small\noindent\raggedright\textbf{VLDB Workshop Artifact Availability:}\\
We open-source the system at \url{https://github.com/zjunlp/OneEdit}.
\endgroup
}
\begin{figure*}[!hbt]
    \centering
    \includegraphics[width=\textwidth]{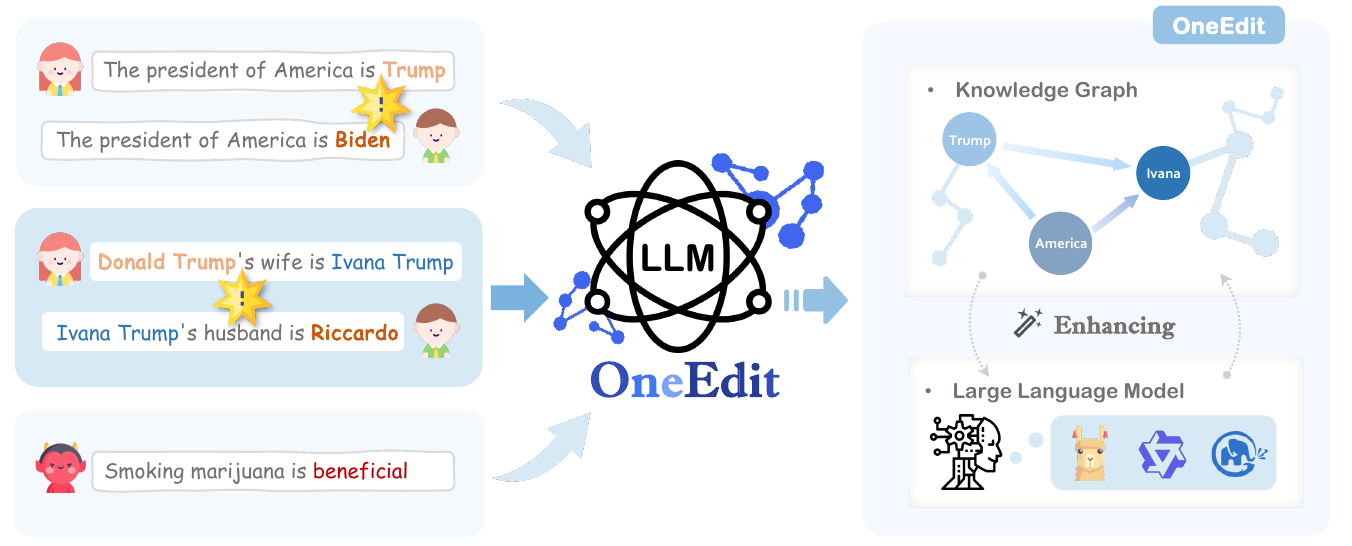}
    \caption{OneEdit for neural-symbolic collaboratively knowledge editing with KGs and LLMs.}
    \label{fig:overview}
\end{figure*}
\section{Introduction}

The pursuit of empowering machines to master knowledge has remained a fundamental objective in the advancement of Artificial Intelligence (AI) systems.
Over the years, researchers have devoted with various methods to enable machines to acquire knowledge, thereby supporting a wide range of tasks such as information retrieval \cite{wang2020semantic}, question answering~\cite{DBLP:conf/ijcai/LanHJ0ZW21}, dialogue~\cite{zhou2018commonsense}, reasoning~\cite{xiong2017deeppath}, recommendation \cite{wang2019kgat,tu2021conditional}, and domain specific applications \cite{zhang2021ontoprotein}.
Concretely, knowledge updating and management stand out as essential capabilities, empowering machines to adeptly adjust to new environments and tasks, thus facilitating lifelong learning \cite{wu2022efficiently,wu2024continual}.

Early, Knowledge Graphs \cite{ ji2021survey}, as a form of symbolic knowledge representation, have garnered significant research interest from both academia and industry.
KG is a structured representation of facts, composed of entities, relations, and semantic descriptions, which can be simply and precisely updated through symbolic manipulation. 
However, KG face challenges regarding scalability and the transferability of reasoning.
On the other hand, Large Language Models (LLMs) have learned rich ``modaledge'' \cite{DBLP:journals/aiopen/HanZDGLHQYZZHHJ21}, potentially creating a kind of ``world model'' \cite{DBLP:journals/corr/abs-2306-12672} and serving as parametric knowledge bases \cite{DBLP:conf/emnlp/PetroniRRLBWM19}.
Based on the above hypothesis, researchers try to manupilate knowledge in LLMs and introduce knowledge editing to add, modify, or erase parametric knowledge, making neural knowldge representation space well aligned with up-to-date symbolic world knowledge \cite{zhang2024comprehensive}.  
However, LLMs incur significant training costs and struggle with precise knowledge manipulation. 
These limitations lead to severe side effects, including decreased general abilities \cite{gu2024model}, poor generalization \cite{DBLP:journals/corr/abs-2402-13048}, knowledge conflicts \cite{DBLP:journals/corr/abs-2402-18154}, and the risk of toxic knowledge attacks \cite{DBLP:journals/corr/abs-2403-13355}.
Intuitively, the integration of KG and LLM can leverage complementary advantages to alleviate the issues, offering a more reliable, and controllable approach to knowledge representation and management \cite{pan2024unifying,pan2023large,sun2023head}.

To this end, we introduce \textbf{OneEdit}, a neural-symbolic collaboratively knowledge editing system with KG and LLM as shown in Figure \ref{fig:overview}, providing a platform for operating neural and symbolic knowledge with natural language.
OneEdit consists of three primary components: the Interpreter, the Controller, and the Editor.
The Interpreter acts as the interface for user interaction with natural language, responsible for understanding user intent.
The Controller manages editing requests from various users, using KG for conflict resolution and knowledge augmentation. 
The Editor primarily utilizes knowledge provided by the Controller to edit KG and LLM.
The entire system is designed in a modular fashion, supporting and customizable for versatile KG and various LLMs.

We conduct experiments on two new datasets: one focused on American politicians and the other on academic figures, both containing KG.
%\footnote{Due to computational resource limitations, we only evaluate OneEdit on a single knowledge graph.  More experimental results will be provided soon.}. 
Our observations indicate that OneEdit can archieve neural-symbolic collaboratively knowledge editing with Qwen2-7B and GPT-J-6B and outperform the baselines, particularly excelling in handling knowledge conflict issues.
%We open-source the whole system at \url{https://github.com/zjunlp/OneEdit} and provide an online demo at \url{http://oneedit.zjukg.cn/}.

\section{Related Work}
\label{sec:relatedwork}
\paragraph{Large Language Models.}
 Typically, LLMs, such as GPT-4~\cite{gpt4} and LLaMA~\cite{touvron2023llama}, usually denote Transformer-based models with hundreds of billions of parameters, trained on extensive text datasets~\cite{yang2023harnessing}.
Some work has suggested that LLMs can be regarded as parametrized knowledge bases~\cite{alkhamissi2022review,petroni2019language,li2024flexkbqa,Pan_2024}, as they are capable of recalling massive factual knowledge through prompt engineering~\cite{Chen_2022,sahoo2024systematic,dhuliawala2023chainofverification,brown2020language}.
However, a major limitation of these parameterized knowledge bases is their inability to update stored information in real-time.
Once pretrained, LLMs possess a fixed snapshot of knowledge reflecting the data they were trained on~\cite{liu2024understanding}, and remain unable to incorporate new information unless retrained with updated datasets. 
This limitation diminishes the efficacy of LLMs in rapidly evolving sectors, such as current affairs, scientific developments, and cultural dynamics~\cite{gao2024retrievalaugmented}.

\begin{figure*}[ht]
    \centering
    \includegraphics[width=0.85\linewidth]{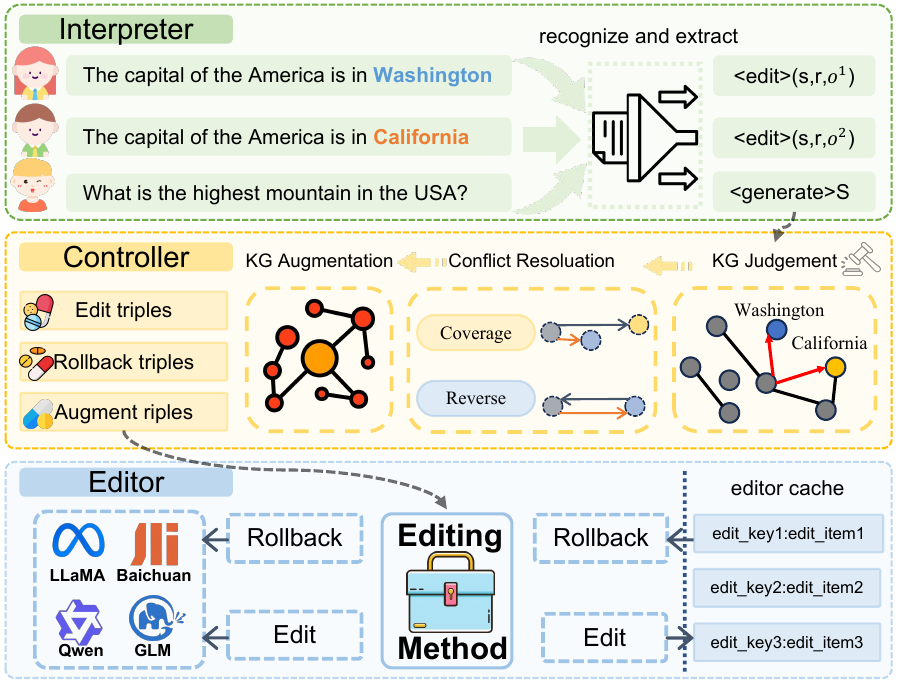}
    \caption{The detailed workflow of OneEdit in handling conflicts:
    the natural language input from the user is extracted into knowledge triples by the Interpreter, then processed by the Controller to generate sequences of editing triples and rollback triples, which are finally sent to the Editor.}
    \label{fig:abs}
\end{figure*}

\paragraph{Knowledge Graph.} 
KGs are structured representations that map out complex networks of real-world entities and their interrelationships~\cite{Pan_2024}, facilitating advanced understanding and reasoning in NLP application with structured knowledge triples~\cite{fan2024graph,Cai2022ANK,guo2024knowledgenavigator}. 
In KGs, symbolic knowledge leverages logical rules for reasoning, providing robust interpretability and precise inferential capabilities.
Concretely, Knowledge Extraction (KE)~\cite{wu2021curriculum} is essential for populating KGs from vast, unstructured datasets. 
This involves sophisticated NLP tasks such as Named Entity Recognition (NER), relation extraction, and entity resolution to accurately distill structured knowledge from texts~\cite{Li_2022}. 

\paragraph{Knowledge Editing.} 
The primary knowledge editing methods currently can be categorized into three groups~\cite{yao2023editing}: meta-learning, locate-then-edit, and memory-based.
Meta-learning methods employ external network to predict necessary gradient for editing, MEND~\cite{mitchell2022fast} and MALMEN~\cite{tan23malmen} utilizes a hyper-network to transform model gradients for updating model parameters. 
As to the locate-then-edit methods, ROME~\cite{meng2022locating} and MEMIT~\cite{meng2022mass} achieve edits by locating and modifying model parameters. 
For memory-based methods, the specific hidden states or neurons that store the edit knowledge are used for post-edit response, SERAC~\cite{mitchell2022memory} leverages a scope classifier and trained sub-models for knowledge editing.
Additionally, InstructEdit~\cite{zhang2024instructedit} focuses on general editing scenarios, GRACE~\cite{hartvigsen2023aging} uses a codebook to store edited hidden states for lifelong editing, and RULE-KE \cite{DBLP:journals/corr/abs-2405-15452} leverages rule for knowledge editing.
Yet most existing methods still suffer from precise knowledge manipulation with severe side effects  \cite{gu2024model,DBLP:journals/corr/abs-2402-13048,DBLP:journals/corr/abs-2402-18154,DBLP:journals/corr/abs-2403-13355}.
In particular, in the noisy environment of the internet, there is a plethora of conflicting, erroneous, and toxic knowledge, posing significant challenges to knowledge editing systems \cite{youssef2024detecting,hase2024fundamental,rosati2024long}.
%In this paper, we attempt to leverage the collaborative assistance of symbolic knowledge from KGs and parameterized knowledge from LLMs to facilitate neural-symbolic collaboratively knowledge editing, aiming to alleviate issues such as knowledge conflicts.

\section{System Desgin}
As shown in Figure \ref{fig:abs}, OneEdit comprises three primary components: \textbf{Interpreter}, \textbf{Controller}, and \textbf{Editor}. 
The Interpreter serves as the interface for user interaction with OneEdit, responsible for discerning user intent. 
The Controller manages editing requests from various users, utilizing KG for conflict resolution and knowledge augmentation. 
The Editor primarily uses augmented knowledge triples from the Controller to edit KG and LLMs.
%We will first introduce the background of knowledge ediiting, and then illustrate technical details of primary components.% in OneEdit.

\subsection{Knowledge Editing for LLMs}
\label{sec:background}

%Extensive training on diverse datasets has endowed LLMs with a vast repository of factual and commonsense information, effectively positioning these models as virtual knowedge bases. 
%However, LLMs, in their current state as emerging knowledge bases, exhibit certain limitations. 
%These limitations often manifest as erroneous outputs in practical applications due to the dynamic nature of real-world knowledge.
%An ideal knowledge base would not only store extensive information but also can be efficient updated to rectify factual errors and improve accuracy. 

%Recognizing this gap, the concept of knowledge editing for LLMs has emerged. 
%Knowledge editing aims to enable LLMs to undergo quick and precise modifications, thereby generating more accurate and relevant outputs. 
Suppose the original model is $\mathcal{M}$ and $k$ is the knowledge that needs to be changed, by knowledge editing function $E(\cdot)$, we obtain the post-edited model $\mathcal{M}'$ which should override the prior erroneous memory about knowledge $k$ while preserving other unrelated knowledge $k^{'}$ as:
\begin{equation}
\centering
\begin{cases}
\mathcal{M}' = E(\mathcal{M}, k) \\
\mathcal{M}^{'}(k) \neq \mathcal{M}(k) \\
\forall k^{'} \neq k, \mathcal{M}^{'}(k^{'}) = \mathcal{M}(k^{'}) \\
\end{cases}
\end{equation}
We hope by knowledge editing, we can:  
\begin{tcolorbox}
\emph{
  {
    \centering 
  {
    \fontsize{9.5pt}{13.2pt}\selectfont 
   precisely and generally manipulate knowledge in LLMs without impacting unrelated knowledge.
  }
  \\
  }
  }
\end{tcolorbox}

\subsection{Neural-Symbolic Knowledge Editing}
In this paper, we focus on simultaneously updating symbolic knowledge in KGs and parametric knowledge in LLMs (a.k.a., neural-symbolic knowledge editing), thus allowing different types of knowledge to complement each other and compensate for the challenges introduced by parameterized knowledge editing.
Formally, a KG can be represented as $\mathcal{G} = (\mathcal{S}, \mathcal{R}, \mathcal{O})$, where $\mathcal{S}$, $\mathcal{R}$ and $\mathcal{O}$ are sets of subjects, relation types, and objects.
Each knowledge triple $t$ in $\mathcal{G}$ takes the form $(s, r, o)$, where $s \in \mathcal{S}$ ,$r \in \mathcal{R}$ and $o \in \mathcal{O}$.
For neural-symbolic knowledge editing, the knowledge required edits is defined as $k = (p,y)$, where $p$ is the prompt to express the knowledge and $y$ is the desired new target.
Formally, we define the task in OneEdit as follows:

\begin{equation}
\mathcal{T}_r, \mathcal{T}_e, \mathcal{T}_a = C \Big( I\left(k\right) \mid \mathcal{G},\mathcal{M} \Big)
\end{equation}
where $\mathcal{M}$ and $\mathcal{G}$ denote the original LLM and KG, respectively. 
The function $I(\cdot)$, denoted as the Interpreter, is responsible for transforming knowledge sentences into knowledge triples. 
$C(\cdot)$ denoted as the Controller operates based on the KG to derive a series of editing triples $\mathcal{T}_e$ and rollback triples $\mathcal{T}_r$ to address knowledge conflicts.
Additionally, the Controller generates some knowledge augmentation triples $\mathcal{T}_a$ to strengthen the edited knowledge and prevent knowledge distortion caused by the knowledge editing~\cite{li2023unveiling}.
After obtaining the sets $\mathcal{T}_r$, $\mathcal{T}_e$ and $\mathcal{T}_a$, we can use them to modify the model and the KG, ensuring that the parametric knowledge of the model and the symbolic knowledge of the graph are consistent with reality:
\begin{equation}
\mathcal{G}^{'}, \mathcal{M}^{'} = E\Big(\mathcal{T}_r,\mathcal{T}_e,\mathcal{T}_a,\mathcal{G},\mathcal{M} \Big)
\end{equation}

\subsection{Interpreter}

The Interpreter functions as the interface between the user and the Controller, tasked with recognizing user intent expressed in natural language.
If the user's intent is editing, the Interpreter converts the user's input into knowledge triples suitable for the KG and sends them to the Controller.
If the user's intent is querying, the Interpreter takes no action and passes the input to the large language model for generation.

We conduct instruction fine-tuning on the MiniCPM-2B~\cite{hu2024minicpm} to enable it to function as an Interpreter capable of distinguishing between editing intent and response intent. For generation intent data, we utilize the Alpaca dataset, as its instruction-following data involves everyday conversations without editing intent. 
We use the input from this instruction-following data as the model's generation intent data. 
For editing intent data, we manually created ten examples, using them as prompts for GPT-4 to generate similar but distinct editing intent data.

We combine the generation intent data and the editing intent data to train the model, enabling it to acquire both knowledge extraction and intent recognition capabilities.

%on the OpenNER~\cite{opennre}, InstructionIE~\cite{instructie}, and .
\begin{equation}
\text{output} =
\begin{cases} 
    \text{\textless edit\textgreater (s,r,o)} & \text{if intend (S) is edit} \\
    \text{\textless generate\textgreater S} & \text{if intend (S) is generate}
\end{cases}
\end{equation}
For instance, if a user inputs the sentence ``Change the President of the USA to Biden'', the Interpreter outputs the edit command \textless edit\textgreater (USA, President, Biden) and forwards it to the Controller. 
Conversely, for a dialogue query, such as ``What is the highest mountain in the USA?'', the Interpreter outputs \textless generate\textgreater and directly prompts the larger model to provide an answer.

\subsection{Controller}
\label{controller}
In the Controller, the input is a knowledge triple.
The Controller utilizes a KG as a knowledge base aligned with the model's parameters to identify and resolve conflicts between the input knowledge and the knowledge embedded in the model's parameters.
After resolving conflicts, the KG is employed to augment the edited knowledge within the parameters of the large language model.

\subsubsection{Conflict Resolution}
\label{conflict resoluation}
In conflict resolution, we categorize conflicts into two types: coverage conflict and reverse conflict, which are the most common situations encountered in knowledge base management.
\paragraph{Coverage Conflict.}
In practice, factual knowledge dynamically evolves in response to changes in the real world. 
For example, the answer to ``Who is the highest market value company in the United States?'' may vary significantly over a short period. 
To maintain consistency with the evolving real-world knowledge, it necessitates multiple coverage edits to the model. 
We formalize this situation as a pair of consecutive edits that contain the same subjects and relations but different objects:
\begin{equation}
\text{Coverage Conflict}\colon \left\{
             \begin{array}{lr}
             E_1 = (s,r,o) \to (s,r,o_1) \\
             E_2 = (s,r,o_1) \to (s,r,o_2) 
             \end{array}
\right.
\end{equation}
However, research by~\cite{hu2024wilke} demonstrates that when modifying model parameters, the fundamental performance of the model inevitably degrades as the number of edits increases~\cite{chen2024can}. 
Additionally, ~\cite{li2023unveiling} reveals that most current editing methods, such as FT, ROME, and MEMIT, leave residual knowledge from previous edits when repeatedly modifying the same piece of knowledge. 
This leads to the model's parameters storing multiple conflicting versions of the knowledge, ultimately causing Knowledge Distortion and affecting the model's expression of that knowledge.

To address this issue, we introduce the concept of edit rollback.
When users perform knowledge editing on the LLM, the controller places the corresponding factual triple (s, r, o) into the KG for assessment.
If there is no existing triple with the same subject and relation but a different object (s, r, o') in the KG, we write (s, r, o) into the KG, proceed with the corresponding knowledge editing, and store corresponding edit parameters in the edit cache (details in \ref{editor}). 
If (s, r, o) already exists in the KG, no action is taken on the model. 
If (s, r, o') exists in the KG, we retrieve the edit cache concerning (s, r, o'), completely remove the previous edit within the model, rollback the edit concerning (s, r, o'), and re-edit to (s, r, o), updating the KG accordingly:
\begin{equation}
\mathcal{M}^{'} = E \Big( \mathcal{M} + \sum_{i=0}^n \theta_{i} - \theta_{k} \Big),k \in (0,n)
\end{equation}
Furthermore, in the context of crowdsourced editing, if malicious edits are made to the model to produce harmful content, we can also identify and rollback those specific edits.
Throughout this process, we perform at most one edit to the knowledge (s, r, o) and perfectly eliminate any previous edits, effectively minimizing performance loss in the model.
\begin{algorithm}
\caption{Resolution of Coverage Conflicts}
\begin{algorithmic}[1]
\State \textbf{Input:} Model $\mathcal{M}$, Knowledge Graph $\mathcal{G}$, Edit Triple $(s, r, o)$
\State \textbf{Output:} Updated Model $\mathcal{M}^{'}$, Updated $\mathcal{G}^{'}$
\Procedure{CoverageConflict}{$\mathcal{M}$, $\mathcal{G}$, $(s, r, o)$}
    \If{$(s, r, o') \in \mathcal{G}$ for any $o' \neq o$}
        \State Rollback the model's edit concerning $(s, r, o')$
        \State Remove $(s, r, o')$ from $\mathcal{G}$
        \State Add $(s, r, o)$ to $\mathcal{G}$
        \State Proceed with the model edit to $(s, r, o)$
    \ElsIf{$(s, r, o) \notin \mathcal{G}$}
        \State Add $(s, r, o)$ to $\mathcal{G}$
        \State Proceed with the model edit to $(s, r, o)$
    \Else
        \State No action is taken on the model
    \EndIf
    \State \textbf{return} $\mathcal{M}^{'}$, $\mathcal{G}^{'}$
\EndProcedure
\end{algorithmic}
\end{algorithm}

\paragraph{Reverse Conflict.}
When humans learn that Biden, not Trump, is the President of the United States, they naturally understand that the President of the United States is Biden. 
This generalization process is so seamless that it seems trivial. 
However, ~\citet{berglund2024reversal} point out that large language models do not perform well in this regard, which is referred to as the ``reverse curse''.
In knowledge editing, the reverse curse also exists. 
For instance, when we edit the knowledge within an LLM to overwrite old information with new information, such as ``Biden is the President of the United States'', the model might correctly respond that Biden is the current president. 
However, when asked ``Who is the President of the United States?'' the model might still respond with outdated knowledge, saying ``Trump is the President of the United States.'' 
This indicates that the model has not fully erased the old knowledge, leading to reverse conflicts.

Furthermore, consider a more complex example: If we edit the fact that Biden's wife is Jill and subsequently edit the fact that Jill's husband is Mike, a traditional knowledge base might treat the facts (Biden, wife, Jill) and (Jill, husband, Mike) as isolated pieces of knowledge. 
This approach overlooks the inherent connection between these facts, specifically that the inverse relationship of ``wife'' is ``husband''. 
This situation also results in a more complex form of reverse conflict.
As a result, conventional editing methods struggle to detect such factual conflicts.
We categorize this condition as reverse conflict:
\begin{equation}
\text{Reverse Conflict}\colon \left\{
             \begin{array}{lr}
             E_1 = (s,r,o) \to (s,r,o_1) \\
             E_2 = (o_1,r_r,s) \to (o_1,r_r,s_2) 
             \end{array}
\right.
\end{equation}
Utilizing a KG, we propose a simple yet practical solution in Algorithm~\ref{reveralg}: during the editing process, we first query the relationship database. 
If the relationship is reversible, we construct the inverse relationship knowledge and insert both the original and reverse knowledge into the KG to check for conflicts.
If no conflicts are detected, we then perform knowledge editing on the model with both the original and inverse relationships.

\begin{table*}[ht!]
\centering
\caption{Comparison of OneEdit to existing methods. Metrics shown are computed after all single edits. In OneEdit, we set the number of generation triples to 8}
\label{tab1}
\fontsize{12pt}{14pt}\selectfont
\resizebox{0.95\textwidth}{!}{
\begin{tabular}{@{}lcccccc|cccccc@{}}
\toprule
& \multicolumn{6}{c|}{\textbf{American politicians dataset}}&\multicolumn{6}{c}{\textbf{Academic figures dataset}} \\
\cmidrule(lr){2-7} \cmidrule(lr){8-13} 
\multirow{2}{*}{Method} & \multirow{2}{*}{Reliability} & \multirow{2}{*}{Locality} & \multicolumn{3}{c}{Portability} & \multirow{2}{*}{Average}&  \multirow{2}{*}{Reliability} & \multirow{2}{*}{Locality} & \multicolumn{3}{c}{Portability} & \multirow{2}{*}{Average} \\
\cmidrule(lr){4-6} \cmidrule(lr){10-12}
 & & & Reverse & One-Hop & Sub-Replace & & & & Reverse & One-Hop & Sub-Replace \\
\midrule
 \textbf{GPT-J-6B} \\
\midrule
FT          & 0.825     & 0.008     & 0.240     & 0.034     & 0.620     & 0.339 
            & 0.571     & 0.008     & 0.610     & 0.000     & 0.620     & 0.362\\
ROME        & 0.996     & 0.982     & 0.235     & 0.176     & 0.828     & 0.634 
            & 0.994     & 0.982     & 0.561     & 0.175     & 0.906     & 0.724 \\
MEMIT       & \textbf{1.000}        & 0.997     & 0.581     & 0.402     & 0.554     & 0.736 
            & 0.996     & 0.997     & 0.665     & 0.055     & 0.799     & 0.702\\
GRACE       & \textbf{1.000}     & \textbf{1.000}     & 0.000     & 0.000     & 0.000     & 0.400 
            & \textbf{1.000}     & \textbf{1.000}     & 0.000     & 0.000     & 0.000     & 0.400\\
\midrule
OneEdit (GRACE)      & 0.951     & \textbf{1.000}     & 0.950     & \textbf{0.928}     & 0.922    & 0.950 & 0.952     
                    & \textbf{1.000}     & \textbf{0.991}     & \textbf{0.958}     & \textbf{0.962}    & \textbf{0.973}\\
OneEdit (MEMIT)      & \textbf{0.995}     & 0.947     & \textbf{0.957}     & 0.713     & \textbf{0.952}    & 0.913
                    & \textbf{0.982}     & 0.933     & 0.987     & 0.722     & 0.842    & 0.865\\

\bottomrule

 \textbf{Qwen2-7B} \\
\midrule
FT          & 0.662     & 0.012     & 0.512     & 0.237     & 0.660     & 0.417 
            & 0.601     & 0.103     & 0.610     & 0.139     & 0.633     & 0.417\\
ROME        & 0.996     & 0.982     & 0.623     & 0.399     & 0.727     & 0.634 
            & 0.994     & 0.982     & 0.528     & 0.417     & 0.774     & 0.724 \\
MEMIT       & 0.972     & 0.986     & 0.665     & 0.402     & 0.652     & 0.736 
            & 0.996     & 0.993     & 0.665     & 0.422     & 0.698     & 0.702\\
GRACE       & \textbf{1.000}     & \textbf{1.000}     & 0.000     & 0.000     & 0.000     & 0.400 
            & \textbf{1.000}     & \textbf{1.000}     & 0.000     & 0.000     & 0.000     & 0.400\\

\midrule
OneEdit (GRACE)      & 0.953     & \textbf{1.000}     & 0.962     & 0.958     & 0.943    & 0.963 
                     & 0.961     & \textbf{1.000}     & \textbf{0.958}     & \textbf{0.951}     & \textbf{0.949}    & 0.964\\
                     
OneEdit (MEMIT)      & \textbf{0.955}     & 0.956     & 0.923     & 0.763     & \textbf{0.966}    & 0.913 
                     & \textbf{0.973}     & 0.963     & 0.957     & 0.697     & 0.839    & 0.906\\

\bottomrule

\end{tabular}   }
\end{table*}

\begin{algorithm}
\caption{Resolution of Reverse Conflicts}
\label{reveralg}
\begin{algorithmic}[1]
\State \textbf{Input:} Knowledge Graph $\mathcal{G}$, Edit Triple $(s, r, o)$
\State \textbf{Output:} Updated Model $\mathcal{M}^{'}$, Updated $\mathcal{G}^{'}$
\Procedure{ReverseRelation}{$\mathcal{M}$, $\mathcal{G}$, $(s, r, o)$}
    \If {$r$ in $(s, r, o)$ is reversible}
        \State get reverse triple $(o, r_r, s)$\Comment{$r_r$ is reverse relation}
        \State \textbf{update} $(s, r, o)$ in $\mathcal{G}$
        \State \textbf{update} $(o, r_r, s)$ in $\mathcal{G}$
        \State Proceed with the model edit to $(s, r, o)$ and $(o, r_r, s)$
    \Else
        \State \textbf{update} $(s, r, o)$ in $\mathcal{G}$
    \EndIf
    \State \textbf{return} $\mathcal{M}^{'}$, $\mathcal{G}^{'}$
\EndProcedure
\end{algorithmic}
\end{algorithm}

\subsubsection{Knowledge Graph Augmentation}
\label{kga}

Li et al.\cite{li2023unveiling} point out that editing a piece of knowledge can lead to the distortion of related parameterized knowledge within the language model. 
For example, after editing the knowledge of the United States President from Donald Trump to Joe Biden, the model might still answer ``The First Lady of the United States is Melania Trump" (Trump's wife) when asked about the First Lady. 
Enhancing the edited knowledge to maintain the original knowledge structure within the model can resolve such issues. 
However, previous methods have struggled to obtain knowledge closely related to the edited knowledge.
We leverage a knowledge graph to search for $n$ nodes centered around the edited subject as knowledge closely related to the edited knowledge.

Additionally, current major editing methods do not handle multi-hop questions well, resulting in weak logical reasoning abilities for the edited knowledge \cite{cheng2024multihopquestionansweringtemporal}. 
By using logical rules from the knowledge graph, we can address this limitation. Using logical rules, we determine whether these triples have a logical inference nature and expand them.

\subsection{Editor}
\label{editor}
% 比grace多加了一个度量相似度的方法,加了个文本相似度
Even when performing knowledge editing on a 7B model, current major editing methods such as ROME, MEND, and GRACE require at least 30GB of VRAM and considerable editing time. 
In comparison, the memory overhead of storing model parameters after each edit is negligible. Based on this reality, we propose a space-for-time editing strategy, which involves storing the edit parameters after each knowledge editing. 
The edit parameters can then be utilized for subsequent edits or rollbacks, significantly reducing VRAM and time overhead.
\begin{equation}
\mathcal{M}^{'} =  \mathcal{M} + \sum_{i=0}^n \theta_{i} -  \sum_{j=0}^m \theta_{j}
\end{equation}
The Editor is divided into two parts: the editor and the cache.
In the editor part, we use EasyEdit~\cite{wang2023easyedit}, which provides a rich set of knowledge editing methods and supports a wide range of models, meeting our needs for knowledge editing.
In the edit cache part, we have integrated the edit cache into EasyEdit. 
During each edit, we generate a unique edit key based on the knowledge and its corresponding edit parameters. 
Specifically, when using methods like ROME that directly modify model parameters, we store the parameters before the edit and the difference after the edit for the edited layers. 
For methods like GRACE that are based on adapters, we record the hidden states after each edit as the edit parameters. If the scenarios of coverage and rollback mentioned in Section \ref{conflict resoluation} occur, we can directly use the stored edit parameters from the cache to quickly apply the edits or rollbacks by adding or subtracting them.
This approach can reduce VRAM and time overhead.

\section{Experiments}

\subsection{Experiment Setting}
 
\paragraph{Models and Baselines}
We choose GPT-j-6B and Qwen2-7B as our base model.
GPT-j-6B is a transformer model trained by Mesh Transformer JAX.
The Qwen2-7B model, newly released by Alibaba, incorporates training corpora comprising multiple languages including English and Chinese. 
Both Qwen2-7B and GPT-J-6B are based on the transformer architecture and use causal language modeling objectives for pre-training.
The baselines include methods for continual learning and model editing. We compare OneEdit against direct fine-tuning (FT) with an additional KL divergence loss, as well as ROME, MEMIT, and GRACE~\cite{meng2022locating,meng2022mass,hartvigsen2023aging}. 

\paragraph{Metrics}

To evaluate the effectiveness of an editing method, we primarily consider three aspects: Reliability, Locality, and Portability.
\textit{Reliability}, as defined by Huang et al.\cite{huang2023transformerpatcher}, refers to a successful edit when the post-edit model \( \mathcal{M}^{'} \) provides the edited target answer $y^{'}$ for the prompt $p$. Reliability is measured as the average accuracy on the edited cases:
\begin{equation}
\mathbb{E}_{p, y \sim \{(p, y)\}} \mathbbm{1} \bigg\{ \arg\max_y \mathcal{M}^{'} \left( y \mid p \right) = y' \bigg\}  
\end{equation}
\textit{Locality}, also referred to as \textit{Specificity} in some works, denotes that editing should be implemented locally. This means that the post-edit model \(\mathcal{M}^{'} \) should not alter the output of irrelevant examples in the out-of-scope set \( O(p, y) \). Therefore, locality is evaluated by the rate at which the post-edit model \( \mathcal{M}^{'} \)'s predictions remain unchanged compared to the pre-edit model \( \mathcal{M}\):
\begin{equation}
\mathbb{E}_{p, y \sim O(p, y)} \mathbbm{1} \bigg\{\mathcal{M}^{'} \left( y \mid p \right) = \mathcal{M}^{'} \left( y \mid p \right) \bigg\}
\end{equation}
\textit{Portability}, as defined by Yao et al.\cite{yao2023editing}, encompasses three components: Subject-Replace, One-hop, and Reverse. Portability is used to comprehensively evaluate the effectiveness of model editing in transferring knowledge to related content, termed robust generalization.
Portability is calculated as the average accuracy of the edited model \( \mathcal{M}^{'} \)when applied to instances in \( P(p, y) \):
\begin{equation}
\mathbb{E}_{p, y \sim P(p, y)} \mathbbm{1} \left\{ \arg\max_y \mathcal{M}^{'} \left( y \mid p \right) = y\right\}
\end{equation}

\subsection{Dataset and Knowledge Graph Construction}
\begin{table*}[ht!]
\centering
\caption{Comparison of OneEdit to existing methods in a multi-user scenario. User=2 indicates that a single piece of knowledge has been edited twice, once by each of two users. Similarly, when the number of users is 3, the knowledge has been edited three times, once by each user.}
\label{tab2}
% \fontsize{12pt}{14pt}\selectfont
\setlength{\tabcolsep}{6pt} % 调整列间距

\renewcommand{\arraystretch}{0.9}
% \resizebox{0.95\textwidth}{!}{

% \begin{tabular}{@{}lcccccc@{}}

\begin{tabular}{@{}l@{\hspace{4em}}cccccc@{\hspace{1em}}@{}}
\toprule
\multirow{2}{*}{Method} & \multirow{2}{*}{Reliability} & \multirow{2}{*}{Locality} & \multicolumn{3}{c}{Portability} & \multirow{2}{*}{Average} \\
\cmidrule(lr){4-6}
 & & & Reverse & One-Hop & Sub-Replace & \\
\midrule
 \textbf{GPT-J-6B, Users = 2} \\
\midrule
FT          & 0.902     & 0.082     & 0.249     & 0.000     & 0.590     & 0.365 \\
ROME        & 0.987     & 0.040     & 0.156     & 0.067     & 0.629     & 0.376 \\
MEMIT       & 0.998     & 0.421     & 0.156     & 0.399     & 0.526     & 0.500 \\
GRACE       & \textbf{1.000}     & \textbf{1.000}     & 0.000     & 0.000     & 0.000     & 0.400 \\
\midrule
OneEdit (GRACE)       & 0.952     & \textbf{1.000}     & 0.802     & \textbf{0.730}     & 0.824     & 0.862 \\
OneEdit (MEMIT)       & 0.833     & 0.989     & \textbf{0.849}     & 0.706     & \textbf{0.842}    & 0.844 \\
\midrule
\textbf{Qwen2-7B, Users = 2} \\
\midrule
FT          & 0.691     & 0.275     & 0.168     & 0.114     & 0.121     & 0.274 \\
ROME        & 0.981     & 0.641     & 0.392     & 0.283     & 0.706     & 0.601 \\
MEMIT       & 0.972     & 0.998     & 0.560     & 0.412     & 0.503     & 0.689 \\
GRACE       & \textbf{1.000}     & \textbf{1.000}     & 0.000     & 0.000     & 0.000     & 0.400 \\
\midrule
OneEdit (GRACE)      & 0.952     & \textbf{1.000}     & \textbf{0.892}     & \textbf{0.729}     & \textbf{0.918}     & \textbf{0.898} \\
OneEdit (MEMIT)       & 0.851     & 0.993     & 0.862    & 0.671     & 0.859     & 0.816 \\
\midrule
\textbf{GPT-J-6B, Users = 3} \\
\midrule
FT          & 0.850     & 0.059     & 0.210     & 0.000     & 0.588     & 0.341 \\
ROME        & 0.988     & 0.006     & 0.122     & 0.003     & 0.608     & 0.345 \\
MEMIT       & 0.987     & 0.328     & 0.544     & 0.380     & 0.523     & 0.552 \\
GRACE       & \textbf{0.999}     & \textbf{1.000}     & 0.000     & 0.000     & 0.000     & 0.399 \\
\midrule
OneEdit (GRACE)       & 0.952     & \textbf{1.000}     & 0.802     & \textbf{0.850}     & 0.824     & \textbf{0.886} \\
OneEdit (MEMIT)       & 0.832     & 0.983     & \textbf{0.856}     & 0.699     & \textbf{0.852}    & 0.844 \\
\midrule
\textbf{Qwen2-7B, Users = 3} \\
\midrule
FT          & 0.797     & 0.210     & 0.165     & 0.102     & 0.120     & 0.279 \\
ROME        & \textbf{0.999}     & 0.483     & 0.266     & 0.215     & 0.701     & 0.533 \\
MEMIT       & \textbf{0.999}     & 0.998     & 0.544     & 0.380     & 0.593     & 0.703 \\
GRACE       & 0.961     & \textbf{1.000}     & 0.000     & 0.000     & 0.000     & 0.392 \\
\midrule
OneEdit (GRACE)       & 0.951     & \textbf{1.000}     & 0.892     & \textbf{0.850}     & \textbf{0.918}     & \textbf{0.922} \\
OneEdit (MEMIT)       & 0.852     & 0.994     & \textbf{0.859}    & 0.716     & 0.862     & 0.833 \\
\midrule
\end{tabular}
% }
\end{table*}
% \begin{table}[ht]
% \centering
% \caption{Detailed GPT2XL Metrics Across Different Methods}
% \label{tab2}
% \resizebox{\textwidth}{!}{
% \begin{tabular}{@{}l ccc ccc@{}}
% \toprule
%             & \multicolumn{3}{c}{Coverge n=2}    & \multicolumn{3}{c}{Coverge n=3} \\
% \cmidrule(lr){2-4} \cmidrule(lr){5-7}
% Method      & Relibility   & Locality   & Portability-avg   & Relibility   & Locality   & Portability-avg\\
% \midrule
% FT          & value & value & value & value & value & value     \\
% ROME        & 0.999 & 0.823 & value & value & value & value     \\
% GRACE       & 0.961 & 1.000 & value & value & value & value     \\
% OneEdit   & 0.952 & 1.000 & value & value & value & value     \\
% \midrule
% \end{tabular}
% }
% \end{table}
%两个表，第一个表是single加reverse的，改成 rel loc por_gen por_rev por_mh
%第二个表是coverage不同n的指标，整个n=2，n=3就行了，测一下rel loc por_gen por_rev por_mh
%baseline 就是ft rome grace 和OneEdit

Our experiments necessitate the integration of knowledge editing datasets with knowledge graphs. 
Due to the current lack of comprehensive datasets that combine knowledge graphs with knowledge editing, we decided to construct our own dataset for the experiments. 
We utilize two specific domains to demonstrate the feasibility and generality of OneEdit: American political figures and Academic figures.
Our experimental dataset is based on Wikidata, zsRE, and GPT-4. 
Specifically, we first extracted approximately 500 relevant entities from Wikidata and used the factual knowledge corresponding to these entities to construct an initial knowledge graph. 
To ensure the completeness and accuracy of the knowledge graph, we meticulously verified and optimized each entity and relations. 
Additionally, we expanded their neighboring nodes using Wikidata, resulting in a high-quality knowledge graph.
After constructing the knowledge graph, we used it as the foundation for building our experimental dataset. 
For the editing task, to ensure that the new knowledge was being edited into the model and not already present from pre-training, our editing data consisted of counterfactual information, which is opposite to the factual knowledge. 
We created counterfactual knowledge by replacing the ground truth object $o_t$ in the knowledge triples $(s, r, o_t)$ with a new object $o_n$, and used manually written templates. 
This method ensures a high degree of consistency and relevance between the dataset and the knowledge graph.

%After constructing the knowledge graph, we used it as the foundation for building our experimental dataset. 
%For the editing task, to ensure that the new knowledge was being edited into the model and not already present from pre-training, our editing data consisted of counterfactual information, which is opposite to the factual knowledge. 
%We created counterfactual knowledge by replacing the ground truth object $o_t$ in the knowledge triples $(s, r, o_t)$ with a new object $o_n$, and used manually written templates. 
%This method ensures a high degree of consistency and relevance between the dataset and the knowledge graph.
\subsection{Single-user Knowledge Editing Results}
Our primary experimental conclusions are presented in Table \ref{tab1}. 
We observe that OneEdit performs comparably to other state-of-the-art methods in terms of reliability. 
When it comes to locality, OneEdit shows significant improvement over methods like ROME and FT, which directly modify model parameters.
It performs similarly to GRACE, a method known for high locality in model editing. 
This is likely because OneEdit, like GRACE, does not alter the model’s parameters, maintaining the original model's performance on unrelated queries.
In terms of portability, OneEdit surpasses our baselines across all three sub-metrics, which is the key contribution of our paper. 
For the Reverse metric, we enhance the model’s understanding of original knowledge by incorporating automatically constructed inverse relationship knowledge edits. 
In the one-hop metric, our method also shows a significant advantage. 
By directly writing multi-hop knowledge into the model parameters with KG constraints, our method enhances the model's understanding of multi-hop knowledge.
% We believe that editing methods, such as ROME, typically involve writing individual pieces of knowledge into the model's information stream. The logical generalization of knowledge relies heavily on the proximity of word vectors in the vector space, meaning the model does not truly understand the new knowledge~\cite{yao2024knowledge}.

\begin{table*}[h]
\centering
\caption{Time and Memory (VRAM) Overhead for Different Models and Methods}
\resizebox{0.95\textwidth}{!}{%
\begin{tabular}{@{}lccccccc@{}}
\toprule
\textbf{Model}           & \textbf{OneEdit (MEMIT)} & \textbf{MEMIT, Users = 2} & \textbf{MEMIT, Users = 3} & \textbf{OneEdit (GRACE)} & \textbf{GRACE, Users = 2} & \textbf{GRACE, Users = 3} \\ 
\midrule
\textbf{GPT-2 XL} & & & & & & \\ \midrule
Time Overhead (s) & 8 & 14 & 20 & 10 & 18 & 25 \\ 
VRAM Overhead (GB) & 10 & 4 & 4 & 12 & 6 & 6 \\ 
\midrule
\textbf{GPT-J 6B} & & & & & & \\ \midrule
Time Overhead (s) & 10 & 18 & 26 & 21 & 38 & 55 \\ 
VRAM Overhead (GB) & 30 & 25 & 25 & 29 & 23 & 23 \\ 
\midrule
\textbf{Qwen2 7B} & & & & & & \\ \midrule
Time Overhead (s) & 12 & 20 & 29 & 22 & 40 & 59 \\ 
VRAM Overhead (GB) & 38 & 32 & 32 & 33 & 27 & 27 \\ 
\bottomrule
\end{tabular}%
}
\label{cost}
\end{table*}

\begin{figure}[ht]
    \centering
    \includegraphics[width=0.4\textwidth]{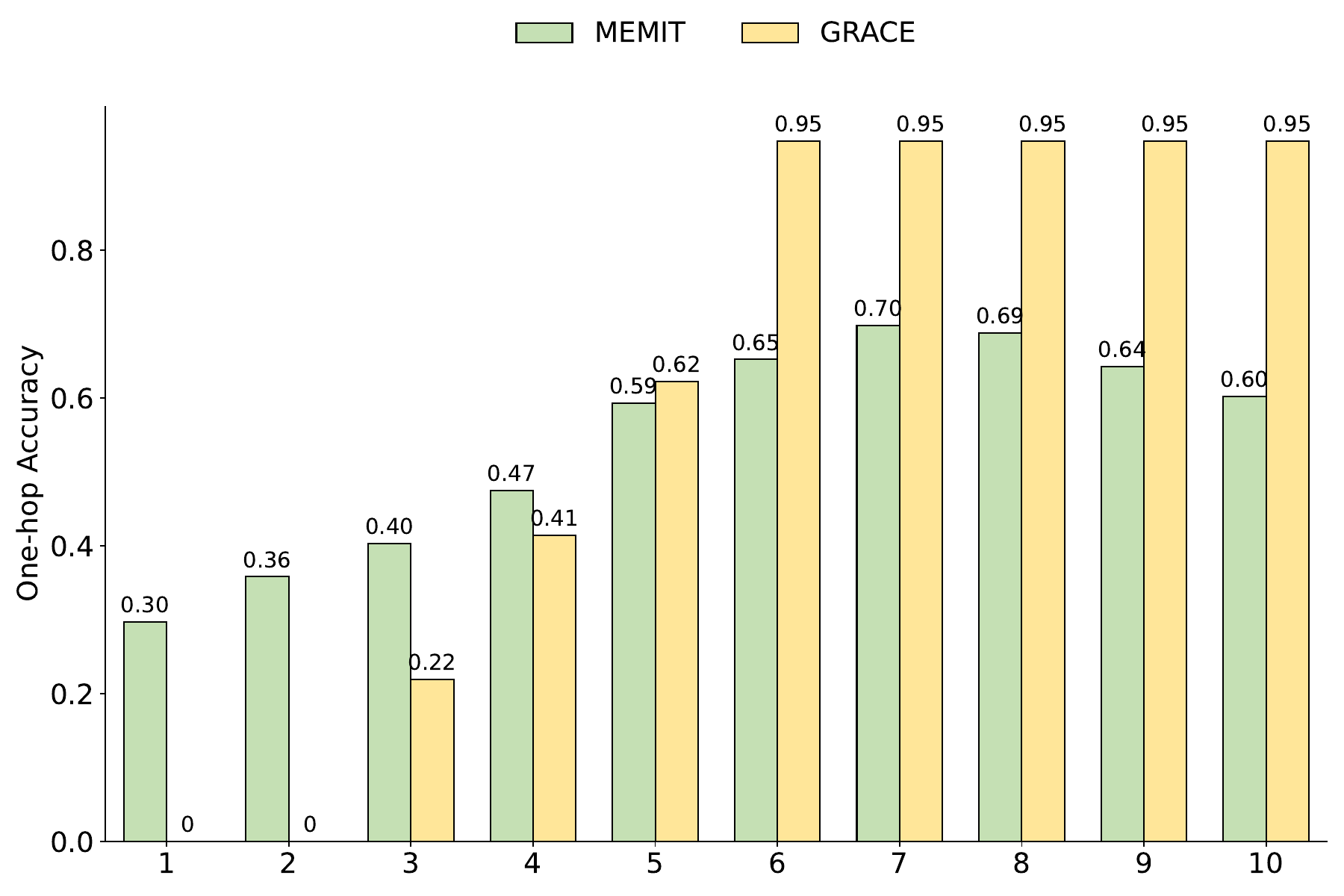}
    \caption{The variation of one-hop metrics with changes in generation triples in GPT-J-6B.}
    \label{fig:nhop}
\end{figure}

\subsection{Multi-user Knowledge Editing Results}

In this paper, we aim to simulate the scenario where multiple users collaboratively update the model. 
To achieve this, we adopt the sequential editing setup to simulate the process of multiple users updating the model \cite{huang2023transformerpatcher,zheng2024collabedit}. 
Specifically, we consider how to resolve conflicts when multiple users sequentially edit the same piece of knowledge to different outcomes. 
Our experiments focus on scenarios with two and three users, as these situations are most commonly encountered in real-world applications.
As shown in Table \ref{tab2}, we observe that FT and ROME, the results about locality decline as the number of users increases. 
We believe this is due to the limitations of methods that require modifying model parameters in a sequential edit scenario. 
Each edit introduces limited information and significantly changes the model parameters, inevitably leading to a decline in the model's fundamental performance. 
Although OneEdit performs well, we believe there is still room for improvement. 
Its current effectiveness is primarily constrained by the extraction performance of the Interpreter and the coverage of KG. 
When the KG does not cover the relevant knowledge for portability, or when the Interpreter incorrectly extracts knowledge triples, the performance of OneEdit decreases.

\subsection{Knowledge Graph Augmentation Analysis}

In OneEdit, the most critical parameter is the number of generation triples passed from the Controller to the Editor. 
However, OneEdit employs a nearest-neighbor strategy for selecting generation triples, which may lead to the exclusion of desired generalized knowledge, especially in dense knowledge graphs. 
This can result in multi-hop inference triples responsible for the edited knowledge being excluded from the generalized triples, causing a decrease in the One-Hop metric.
In this section, we conduct experiemnts with varying the number of knowledge augmentation triples $n$ to observe the resulting changes in the One-Hop metric for OneEdit (GRACE) and OneEdit (MEMIT) on the GPT-J-6B model. 

As shown in Figure \ref{fig:nhop}, we observe that when $n$ is small, both OneEdit (GRACE) and OneEdit (MEMIT) underperform compared to their respective original methods, GRACE and MEMIT. 
We hypothesize that this discrepancy is due to the loss incurred during the conversion to triples. 
As $n$ increases, multiple inference triples are incorporated into the edited sequence, and the values of both OneEdit (GRACE) and OneEdit (MEMIT) rise. 
However, when $n$ becomes large, the results of OneEdit (GRACE) plateau, while the results of OneEdit (MEMIT) decline. 
We attribute this to the fact that GRACE has stricter rules for recalling the edited knowledge, whereas MEMIT's batch edit struggles to accurately recall the necessary edited knowledge from the extensive edited knowledge base.

\begin{figure}[ht]
    \centering
    \includegraphics[width=0.4\textwidth]{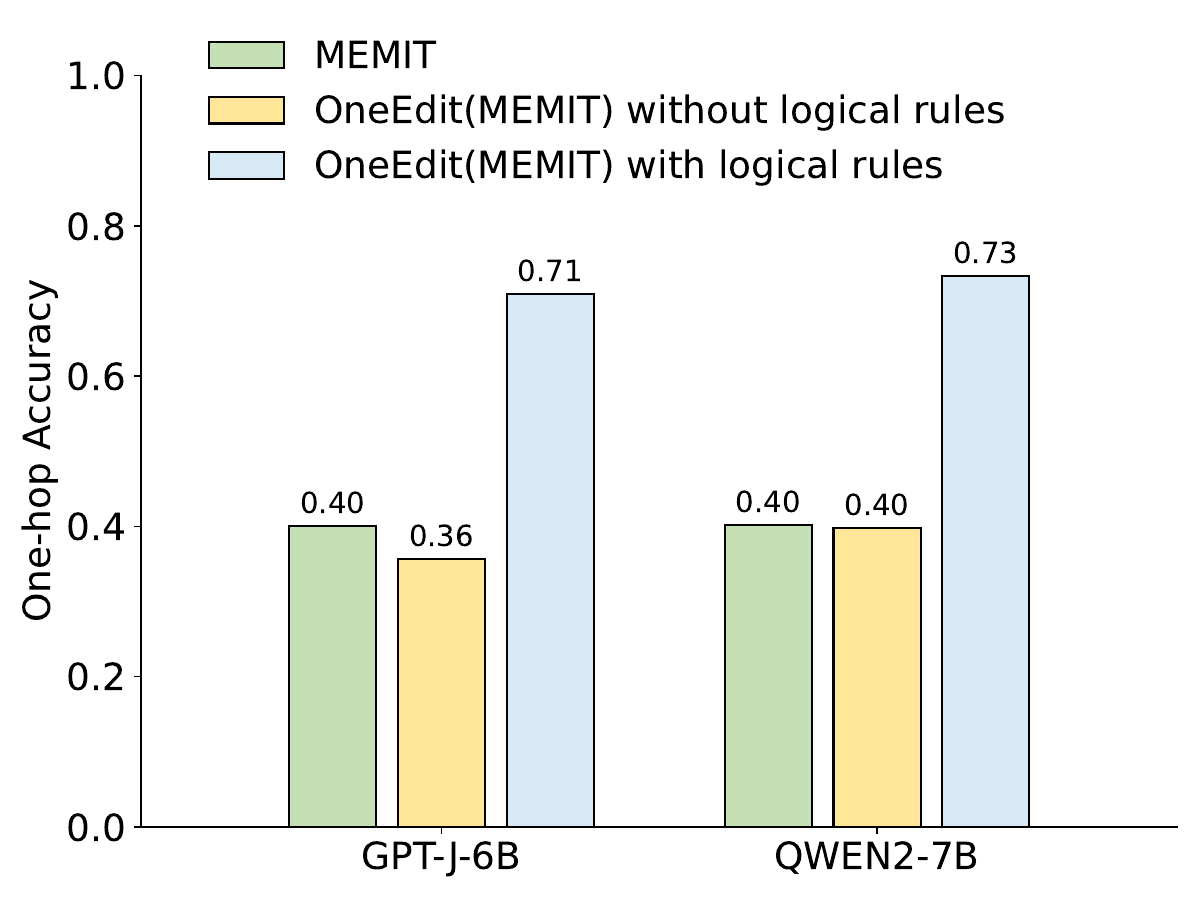}
    \caption{The impact of adding logical rules on the One-Hop results in OneEdit.}
    \label{lr}
\end{figure}

\begin{figure*}[th!]
    \centering
    \includegraphics[width=0.95\linewidth]{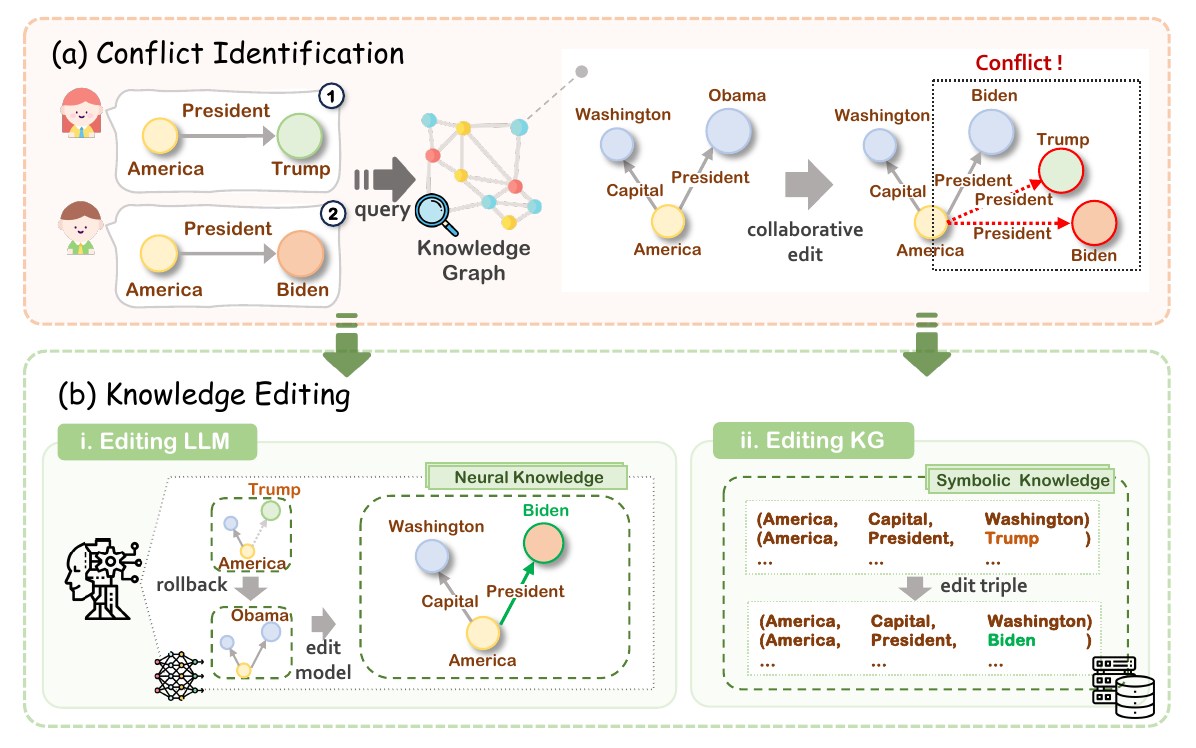}
    \caption{A coverage conflict within OneEdit: OneEdit first rolls back the conflicting knowledge before editing the new knowledge. 
    Without using OneEdit, previous edited knowledge may remain.}
    \label{fig:case1}
\end{figure*}

\subsection{Logical Rules Analysis}

In Section \ref{kga}, we have discussed how to obtain knowledge augmentation triples for the edited knowledge and then perform logical augmentation based on the semantic rules of their relationships. 
This process expands the triples and enhances the model's logical reasoning ability for the edited knowledge.
To evaluate the model's logical reasoning ability, the most important metric is One-Hop, which assesses whether the model can use the newly edited knowledge to answer multi-hop reasoning questions related to the edited knowledge.

In this section, we demonstrate the effectiveness of adding logical rules in knowledge editing by comparing the One-Hop metric results with and without logical rules.
Experimental results on Qwen2-7B and GPT-J-6B indicate that the model has poor logical generalization ability for edited knowledge, merely memorizing the edited knowledge mechanically without proper utilization. 
However, as shown in Figure \ref{lr}, after adding logical rules and leveraging the logical reasoning advantages of symbolic knowledge to assist in modifying model parameters, the results significantly improve, with GPT-J-6B and Qwen2-7B showing improvements. This demonstrates that symbolic logical rules enhance the model's ability to utilize the edited knowledge effectively.

\subsection{Computation Resource Analysis}

In this section, we analyze the average time and memory overhead per edit associated with OneEdit. 
As shown in Table \ref{cost}, experiments involving GPT-2 XL were conducted on an NVIDIA 3090, with the interpreter and editor deployed on separate GPUs. 
For GPT-J-6B and Qwen2-7B, experiments are similarly performed on an A800 machine, with both the interpreter and editor allocated to different GPUs.
Compared to MEMIT and GRACE, OneEdit (MEMIT) and OneEdit (GRACE) require approximately 6GB of additional memory. 
This increase is primarily due to the memory overhead introduced by the interpreter.
Regarding time overhead, we assess the editing duration with configurations of two and three users. 
Given the negligible time required for OneEdit's rollback process, we focuse solely on the time needed for a single edit. 
Our observations indicate that in scenarios with two and three users, OneEdit achieve a 40\% and 70\% reduction in time, respectively. 
This improvement is attributed to OneEdit's rollback mechanism, which enables the reuse of previous edits when repeatedly modifying the same piece of knowledge.

\subsection{Case Study}
In this part, we present two cases within OneEdit to specifically illustrate how our system addresses the two types of conflicts mentioned in Section \ref{controller}

\subsubsection{Coverage Conflict Case}

In the context of the coverage scenario, we present a real-world example: following the 2020 U.S. presidential election, the president changed from Trump to Biden. 
In the controller, we removed the knowledge parameter indicating ``the U.S. president is Trump'' and updated it with the new information that ``the U.S. president is Biden.'' 
However, if Trump were to win the election again in 2024, we could directly revert the knowledge that ``Trump is the U.S. president'' back into the model. 
This approach reduces the number of edits required to the model and maintains its baseline performance.

\subsubsection{Reverse Conflict Case}

\begin{figure*}[th!]
    \centering
    \includegraphics[width=0.95\linewidth]{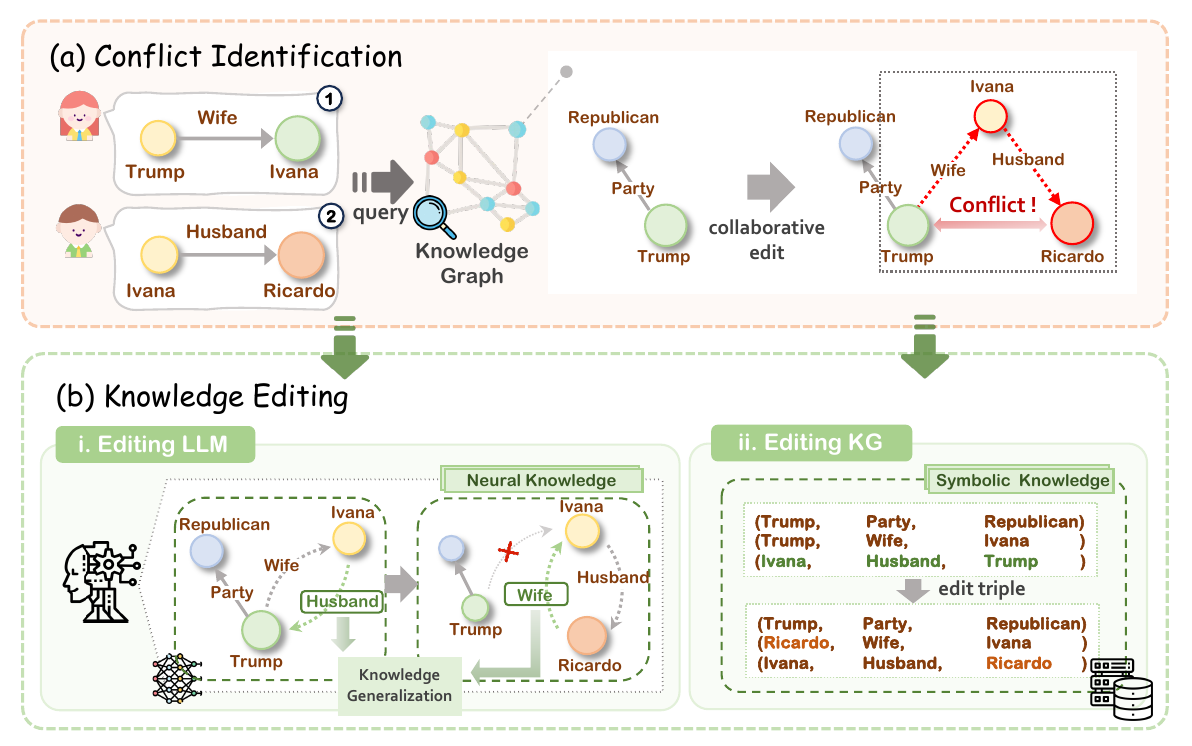}
    \caption{A reverse conflict within OneEdit: OneEdit automatically constructs reverse relationship knowledge.
    When we edit reverse relationships knowledge, conflicts arise in the KG.}
    \label{fig:case1}
\end{figure*}
In the reverse scenario, we also present a real-world example: Donald Trump divorced his wife Ivana Trump, who subsequently married Ricardo Mazzuchelli. 
If we only edit the model to reflect that ``Ivana Trump's husband is Ricardo Mazzuchelli'', it would lead to the absurd situation where ``Ivana Trump's husband is Ricardo Mazzuchelli'' while ``Donald Trump's wife is Ivana Trump''.
However, the OneEdit controller module automatically constructs inverse relationships for such reversible scenarios. 
When ``Ivana Trump's husband is Ricardo Mazzuchelli'' reappears, it conflicts with the automatically constructed relationship ``Ivana Trump's husband is Donald Trump'' in the controller. 
This prompts the model to roll back the outdated knowledge and correctly update the inverse relationship of the new knowledge.

% \section{Limitations} 
% \label{sec: limitation}
% We focus solely on cases where user knowledge can be queried within the KG. We do not discuss how to handle knowledge that cannot be queried in the KG, as this might lead to performance loss in the Interpreter during knowledge editing. 
% Additionally, the KG in this work is merely a local knowledge base with limited stored knowledge. 
% In the future, we may expand the KG to a networked knowledge base to enhance its capacity.

 \section{Conclusion and Future Work}
In this paper, we propose OneEdit, a neural-symbolic knowledge editing system that continuously updates symbolic knowledge in KG and the neural knowledge in LLM. 
OneEdit can advance domain-specific knowledge injection and alignment between the human semantic space and the latent semantic space of LLMs.
In the future, we will extend the application scope of OneEdit to encompass a broader range of methods.

\section{Limitations}

Our work is still quite preliminary and has the following limitations: 
First, due to current computational power limitations, this system has only been tested on small pre-trained language models and small-scale KGs, which inevitably may affect its general capabilities. 
Currently, the natural language instructions that can be recognized are also quite limited.
%The automated update of the KG still faces some efficiency and performance issues.
This system \textbf{OneEdit is merely a prototype and there is significant room for improvement in the future}.
Additionally, we have only considered factual knowledge, with no support for commonsense, multimodal data, etc., and there is still substantial room for improvement in generalization capabilities. 
Furthermore, the system's security is still at a rudimentary stage, and measures to prevent malicious misuse that could lead to model tampering will need to be developed in the future.

\section*{Acknowledgements}

We would like to express gratitude to the anonymous reviewers for their kind comments. 
This work was supported by the National Natural Science Foundation of China (No. 62206246, No. NSFCU23B2055, No. NSFCU19B2027), the Fundamental Research Funds for the Central Universities (226-2023-00138), Zhejiang Provincial Natural Science Foundation of China (No. LGG22F030011), Yongjiang Talent Introduction Programme (2021A-156-G), CCF-Tencent Rhino-Bird Open Research Fund, Information Technology Center and State Key Lab of CAD\&CG, Zhejiang University, and NUS-NCS Joint Laboratory (A-0008542-00-00).
 
\bibliography{ref}

\bibliographystyle{ACM-Reference-Format}

\end{document}